\documentclass[sigconf]{acmart}

\copyrightyear{2025}
\acmYear{2025}
\setcopyright{cc}
\setcctype{by-nc-sa}
\acmConference[MM '25]{Proceedings of the 33rd ACM International Conference on Multimedia}{October 27--31, 2025}{Dublin, Ireland}
\acmBooktitle{Proceedings of the 33rd ACM International Conference on Multimedia (MM '25), October 27--31, 2025, Dublin, Ireland}\acmDOI{10.1145/3746027.3754560}
\acmISBN{979-8-4007-2035-2/2025/10}

\begin{document}

\title{Towards Fine-Grained Human Motion Video Captioning}


\author{Guorui Song}
\authornote{Both authors contributed equally to this research.}
\email{sgr24@mails.tsinghua.edu.cn}
\orcid{0009-0004-5803-5924}
\affiliation{%
  \institution{Tsinghua University}
  \city{Shenzhen}
  \country{China}
}

\author{Guocun Wang}
\authornotemark[1]
\email{20217862@stu.neu.edu.cn}
\orcid{0009-0000-1815-2342}
\affiliation{%
  \institution{Tsinghua University}
  \city{Shenzhen}
  \country{China}
}

\author{Zhe Huang}
\email{huangz23@mails.tsinghua.edu.cn}
\orcid{0009-0007-4722-8749}
\affiliation{%
  \institution{Tsinghua University}
  \city{Shenzhen}
  \country{China}
}

\author{Jing Lin}
\email{jinglin.stu@gmail.com}
\orcid{0009-0006-7549-6874}
\affiliation{%
  \institution{ByteDance}
  \city{Singapore}
  \country{Singapore}
}

\author{Xuefei Zhe}
\email{zhexuefei@outlook.com}
\orcid{0000-0002-5005-7166}
\affiliation{%
  \institution{City University of Hong Kong}
  \city{Hong Kong}
  \country{Hong Kong}
}

\author{Jian Li}
\email{l-j21@mails.tsinghua.edu.cn}
\authornote{Corresponding Author.}
\orcid{0009-0007-0466-5930}
\affiliation{%
  \institution{Tsinghua University}
  \city{Shenzhen}
  \country{China}
}

\author{Haoqian Wang}
\authornotemark[2]
\email{wanghaoqian@tsinghua.edu.cn}
\orcid{0000-0003-2792-8469}
\affiliation{%
  \institution{Tsinghua University}
  \city{Shenzhen}
  \country{China}
}

\renewcommand{\shortauthors}{Guorui Song et al.}
\begin{abstract}
Generating accurate descriptions of human actions in videos remains a challenging task for video captioning models. Existing approaches often struggle to capture fine-grained motion details, resulting in vague or semantically inconsistent captions. In this work, we introduce the Motion-Augmented Caption Model (M-ACM), a novel generative framework that enhances caption quality by incorporating motion-aware decoding. At its core, M-ACM leverages motion representations derived from human mesh recovery to explicitly highlight human body dynamics, thereby reducing hallucinations and improving both semantic fidelity and spatial alignment in the generated captions. To support research in this area, we present the Human Motion Insight (HMI) Dataset, comprising 115K video-description pairs focused on human movement, along with HMI-Bench, a dedicated benchmark for evaluating motion-focused video captioning. Experimental results demonstrate that M-ACM significantly outperforms previous methods in accurately describing complex human motions and subtle temporal variations, setting a new standard for motion-centric video captioning.
\end{abstract}

\ccsdesc[300]{Computing methodologies~Video summarization}
\ccsdesc[300]{Computing methodologies~Natural language generation}

\keywords{Video Captioning, Human Motion, Video Dataset}


\maketitle
\section{Introduction}

The rapid advancement of vision-language model has significantly improved video understanding capabilities across various applications. However, accurately capturing and describing human motion in videos remains a substantial challenge. Current state-of-the-art multimodal models \cite{Improved2024, InternVL2024, LLaMAVID2023, LLaVANeXTInterleave2024, LLaVAOneVision2024, Qwen2VL2024, QwenVL2023, VideoChatGPT2024a, VideoLLaMA2023, VILA2024a} often struggle with precisely describing human movements, frequently generating hallucinations, providing overly generic descriptions, or incorrectly identifying body parts and actions. This limitation is particularly evident when models are tasked with describing fine-grained movements in videos, such as human body movements, subtle facial expressions, and other nuanced physical actions.

Existing open-source multimodal large language models (e.g., Video-LLaVA\cite{videollava}, LLaVA-OV\cite{LLaVAOneVision2024}, Qwen2VL\cite{Qwen2VL2024} ) demonstrate notably poor performance in human motion captioning tasks. They tend to produce vague or inaccurate answers and struggle to capture the fine-grained details of human body movements. For instance, when presented with a yoga video, these models might simply state that "a person is doing yoga" without describing the specific pose, body positioning, or movement. In the domain of motion-to-text conversion, most approaches rely on simplistic text combination methods without effectively leveraging the capabilities of large language models, resulting in limited generalization ability.

Even specialized motion models\cite{arbolBodyShapeGPTSMPLBody2024,bugliarelloWhatAreYou2025,pangGlobalPositionAware2025,wangFgT2MLLMsAugmentedFineGrained2025,zhaoEfficientMotionAwareVideo2025,chen2024motionllm} integrated with large language models, such as MotionLLM\cite{chen2024motionllm}, exhibit significant limitations when applied to video tasks. These models typically generate brief, overgeneralized responses and still suffer from severe hallucination issues. They demonstrate weak perception of micro-expressions, background information, and task states. Moreover, their inability to accurately describe complex body positions and movements severely restricts their practical utility in real-world applications requiring detailed human motion understanding.

\begin{figure*}[htbp] 
    \centering
    \includegraphics[width=\textwidth]{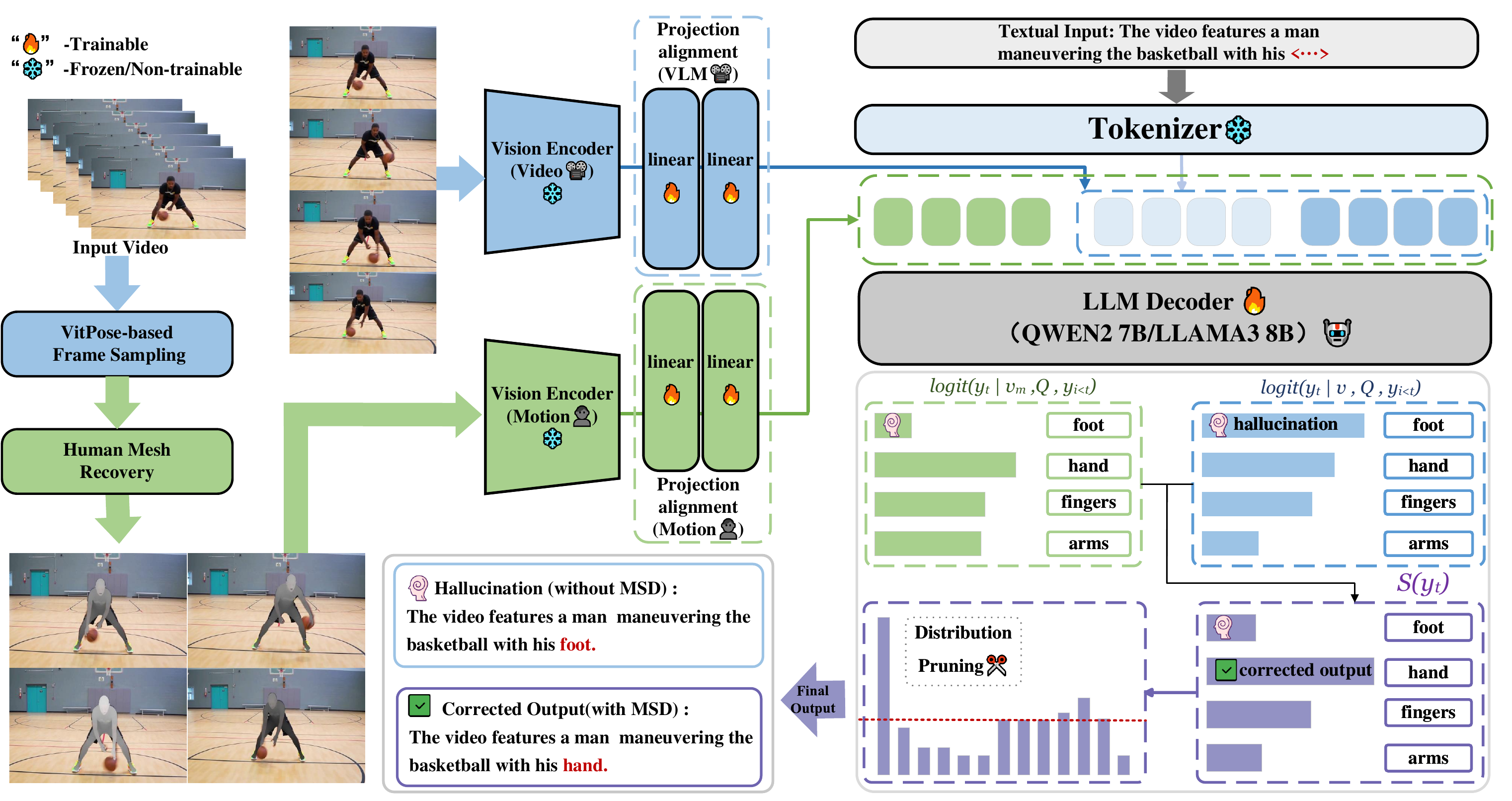} 
    \caption{Overview of our proposed M-ACM (Motion-Augmented Caption Model) framework. The system processes input videos through dual pathways: a standard visual pathway (top) and a motion-specialized pathway (bottom). The visual pathway extracts general visual features via a frozen vision encoder, while the motion pathway uses ViTPose-based ~\cite{xu2022vitpose} frame sampling and human mesh recovery to generate precise motion representations. Both representations are projected into a common embedding space through trainable projection alignment modules. Our key innovation, Motion Synergetic Decoding (MSD), addresses hallucination issues by comparing logit distributions from both pathways. As shown in the example, without MSD the model incorrectly identifies the basketball being handled with the "foot" (hallucination), whereas with MSD the model correctly identifies the "hand" as the body part manipulating the ball.}
    \label{pipeline}
\end{figure*}

 The dataset limitations further compound these problems, as motion domain datasets are predominantly synthetic collections \cite{humanml3d,Plappert_2016,AMASS:ICCV:2019,delmas2022posescript,delmas2023posefix}, which diverge significantly from real-world videos, lacking complex background information and interactions between humans and environments. Existing open-source video datasets\cite{chen2024motionllm,grauman2024egoexo4dunderstandingskilledhuman,hong2024motionbench,zhang2024videoinstructiontuningsynthetic,ju2024miradatalargescalevideodataset,chen2024we,chen2024sharegpt4video} mostly contain general descriptions without specific annotations for fine-grained human action understanding. 
 
 Current approaches to mitigating hallucinations\cite{chuangDoLaDecodingContrasting2024,heTopiclevelSelfCorrectionalApproach2024,dengWordsVisionVisionLanguage2025,Mitigating2025b,yinWoodpeckerHallucinationCorrection2023}, such as OPERA\cite{huangOPERAAlleviatingHallucination2024} and VCD\cite{
Mitigating2024a}, primarily focus on computationally expensive multiple generation/query iterations without incorporating additional prior information, ultimately failing to address the fundamental issue of inadequate motion representation.

In summary, our main contributions are as follows:

• Motion Synergetic Decoding (MSD) Framework: A novel approach that reduces hallucination in human motion captioning by combining standard visual understanding with specialized motion representations from SMPL-based human mesh recovery. The dual-pathway system with synergy calculation mechanism significantly improves motion description accuracy.

• Human Motion Insight (HMI) Dataset: A comprehensive resource with 115K videos and descriptions focused on human movement, sourced from five diverse datasets. Created through rigorous filtering, motion quality assessment, and ViTPose-based dynamic sampling to facilitate human motion understanding.

• HMI-Bench: A specialized benchmark for evaluating motion caption generation across multiple dimensions including detailed movement descriptions, spatial relationships, temporal dynamics, and semantic accuracy. Experiments show the framework outperforms existing methods on both standard and specialized metrics.

\section{Related Work}

\subsection{Video Understanding with Large Language Models}

Video understanding has emerged as a pivotal research domain in contemporary academia, as it aligns more closely with human perceptual mechanisms for processing real-world information. With the rapid development of multimodal large models, an increasing number of works\cite{Improved2024, InternVL2024, LLaMAVID2023, LLaVANeXTInterleave2024, LLaVAOneVision2024, Qwen2VL2024, QwenVL2023, VideoChatGPT2024a, VideoLLaMA2023, VILA2024a} have demonstrated remarkable capabilities in video understanding. 

Early works\cite{VideoChatGPT2024a, VideoLLaMA2023,LLaMAVID2023} connect LLM and pretrained vision encoders like CLIP\cite{Learning2021} through a projection module and finetune the projection module with video instruct data. Towards advancing the development of multimodal large models, VILA\cite{VILA2024a} conducts systematic research on pre-training vision-language models, establishing a methodological framework that provides significant insights for subsequent multimodal large model training. Recent advancements\cite{LLaVANeXTInterleave2024, LLaVAOneVision2024, Qwen2VL2024, InternVL2024} have re-engineered the visual representations and visual encoder architecture to accommodate diverse visual input modalities and enhance model perceptual capabilities. They also strategically curated training data and further enabled substantial performance improvements in multimodal large models.

However, these methods primarily focus on video understanding for general purpose, resulting in vague or inaccurate answers in human motion understanding.

\subsection{Video Understatnding Datasets}
An increasing number of well-curated text-video datasets\cite{chen2024motionllm,grauman2024egoexo4dunderstandingskilledhuman,hong2024motionbench,zhang2024videoinstructiontuningsynthetic,ju2024miradatalargescalevideodataset,chen2024sharegpt4video, zhang2025motion, lin2023motionx} have emerged as valuable resources for video understanding. However, most of them\cite{grauman2024egoexo4dunderstandingskilledhuman,zhang2024videoinstructiontuningsynthetic,ju2024miradatalargescalevideodataset,chen2024sharegpt4video} are for general purpose and demonstrate limited capacity for representing fine-grained motion patterns. Building upon 
the similar goal to enhance human motion understanding in video, some other works\cite{chen2024motionllm, zhang2025motion, lin2023motionx, hong2024motionbench} curate video annotation specializing in human bodily motion analysis. But they still lack fine-grained motion description.

\subsection{Hallucination Mitigation in Multimodal Large Language Models}
Hallucinations in MLLMs, which means generating factually inconsistent content, pose critical reliability challenges in practical applications. To mitigate hallucinations, efforts have been made through data quality enhancement\cite{chen2025perturbollava, Reflective2024Zhang}, model architecture refinement\cite{he2024incorporatingvisualexpertsresolve, Eyes2024Tong}, training strategy optimization\cite{Hallucination2024Jiang, chen2023mitigating}, and inference process redesign\cite{Look2024, Mitigating2024a, Mitigating2025b}.
In this work, we focus on methods that mitigate hallucination in inference stage. As a representative approach, contrastive decoding\cite{Mitigating2024a, Mitigating2025b} modulates the generated probability distributions during the inference phase to mitigate hallucination. 

\section{Proposed Method}

\subsection{System Overview}

As shown in Figure~\ref{pipeline}, our proposed approach addresses hallucination issues in motion caption generation by incorporating Motion Synergetic Decoding(MSD). The system takes video input and processes it through two parallel pathways: a standard visual understanding pathway and a motion-specialized pathway.

In the standard pathway, the input video is processed by a vision encoder to extract visual features. Concurrently, in the motion pathway, the same input video undergoes ViTPose-based ~\cite{xu2022vitpose} frame sampling followed by human mesh recovery to generate motion-specific representations that are then encoded by a motion-specialized vision encoder.

Both visual and motion-based representations are projected into a common embedding space through dedicated projection alignment modules. These aligned representations are then processed by an LLM decoder to generate the final output. The key innovation of our approach is the synergy calculation between these two modalities, which effectively mitigates hallucination issues by leveraging complementary information from both pathways.

\subsection{Motion-Aware LLM Finetuning}

For the motion caption generation task, we formulate the problem as a conditional sequence generation task. Given a video input containing human motion, the model generates descriptive text capturing the precise actions and movements. 

The generation probability at each time step can be expressed as: 
\begin{equation} 
\begin{aligned} 
p(y_{T}|v, Q) = \prod_{t=1}^{T} p(y_t|v, Q, y_{<t}) ,
\end{aligned} 
\end{equation} 

where $v$ represents the visual input, $Q$ denotes the instruction or query, and $y_t$ is the token generated at timestep $t$, with $y_{<t}$ representing all previously generated tokens.

To overcome standard vision-language models' struggles with accurately interpreting human motion, our approach enhances vision-language models with motion-specific information. We fine-tuned a 2-layer MLP projector and LLM decoder on our Human Motion Insight (HMI) dataset while keeping the vision encoder frozen, enabling specialized motion understanding without compromising general reasoning capabilities.

\subsection{Human Motion Representation}



Despite our fine-tuning approach, the model still exhibited hallucination issues when identifying and describing human body parts and movements. To address this, we incorporate the NLF\cite{sarandi2024nlf} method, which leverages SMPL-based\cite{loper2023smpl} human mesh recovery to generate precise body representations from video frames. This approach allows us to create motion-specific masks $v_m$ that augment the standard visual input $v$ in our generation formula: 

\begin{equation} \begin{aligned} 
p(y_{T}|v, v_m, Q) = \prod_{t=1}^{T} p(y_t|v, v_m, Q, y_{<t}) .
\end{aligned} \end{equation}

Our motion representation extraction process involves ViTPose-based ~\cite{xu2022vitpose} frame sampling, NLF-inspired \cite{sarandi2024nlf}  localizer functions for precise body point localization, SMPL \cite{loper2023smpl} model fitting to create consistent human mesh representations, and generation of motion-specific masks highlighting relevant body regions.

By generating detailed motion masks that highlight specific body regions and their trajectories, we provide complementary information that helps disambiguate similar body parts and accurately track fine-grained movements, particularly valuable for articulated structures like fingers and hands. The motion representation serves as a specialized attention guide, directing the model's focus to relevant body regions during generation, reducing reliance on potentially ambiguous visual features.




\subsection{Motion Synergetic Decoding}

Our core objective is to enhance the model's understanding by capturing the complementary information between these modalities. To achieve this, we construct a basic synergy term:
\begin{equation}
\begin{aligned}
\mathcal{S}_{\text{basic}}(y_T) = & \frac{1}{2} \sum_{t=1}^{T} \left( logit(y_t \mid v, Q, y_{<t}) + logit(y_t \mid v_m, Q, y_{<t}) \right) \\
& + \gamma \cdot \min \left( logit(y_t \mid v, Q, y_{<t}), logit(y_t \mid v_m, Q, y_{<t}) \right).
\end{aligned}
\end{equation}

This formula combines average logits and their minimum value (weighted by $\gamma$) to balance influence.

The first term represents the arithmetic mean of the logits from both modalities, ensuring equal contribution from each source.The term prevents any single modality from dominating the prediction process, creating a foundation for robust multimodal understanding. The second term, controlled by parameter $\gamma$, emphasizes consensus between modalities by focusing on the minimum logit values, which helps prevent overconfidence when one modality produces misleading predictions. By prioritizing agreement between information sources, this mechanism serves as a crucial safeguard against potential errors or noise in individual modalities.Due to limitations in capturing complex relationships, we developed a comprehensive formula with five components:

\begin{equation}
\begin{aligned}
\mathcal{L}_1(y_T) = \alpha_1 \cdot \sum_{t=1}^{T} \exp\left( logit(y_t \mid v, Q, y_{<t}) \right)\\
+ \alpha_2 \cdot \sum_{t=1}^{T} \exp\left( logit(y_t \mid v_m, Q, y_{<t}) \right).
\end{aligned}
\end{equation}

The first component $\mathcal{L}_1$ amplifies high-confidence predictions through exponential terms. This exponential transformation magnifies the contribution of tokens with higher logit values, and $\alpha_1$ and $\alpha_2$ control the relative importance of each modality, enabling flexible adjustment based on their respective reliability.

\begin{equation}
\begin{aligned}
\mathcal{L}_2(y_T) = \alpha_3 \cdot \sum_{t=1}^{T} \left( \frac{logit(y_t \mid v, Q, y_{<t}) + logit(y_t \mid v_m, Q, y_{<t})}{2} \right).
\end{aligned}
\end{equation}

The second component $\mathcal{L}_2$ maintains balanced contributions as a stabilizing factor. By computing the arithmetic mean of logits from both modalities, $\mathcal{L}_2$  ensures that neither source dominates the prediction process. The parameter $\alpha_3$ adjusts the overall influence of this averaging mechanism in the final synergy score.

\begin{equation}
\begin{aligned}
\mathcal{L}_3(y_T) = \alpha_4 \cdot \sum_{t=1}^{T} 
    \log\biggl( 1 + \Bigl| logit(y_t \mid v, Q, y_{<t}) 
    & \\
    - logit(y_t \mid v_m, Q, y_{<t}) \Bigr| \biggr).
\end{aligned}
\end{equation}


The third component $\mathcal{L}_3$ introduces a soft penalty for modal disagreements using logarithmic scaling. $\mathcal{L}_3$ increases as the absolute difference between modality logits grows, effectively discouraging predictions where the two sources strongly disagree. 

\begin{equation}
\begin{aligned}
\mathcal{L}_4(y_T) = \alpha_5 \cdot \sum_{t=1}^{T} \left( \left( logit(y_t \mid v, Q, y_{<t}) - logit(y_t \mid v_m, Q, y_{<t}) \right)^{\theta} \right).
\end{aligned}
\end{equation}

The fourth component $\mathcal{L}_4$ captures non-linear relationships through power transformation with parameter $\theta$. By adjusting $\theta$, we can control the sensitivity to various degrees of modal agreement or disagreement, enhancing the model's ability to handle complex interactions.

\begin{equation}
\begin{aligned}
\mathcal{L}_5(y_T) = \alpha_6 \cdot \sum_{t=1}^{T} \left( \left| logit(y_t \mid v, Q, y_{<t}) - logit(y_t \mid v_m, Q, y_{<t}) \right| \right)^2.
\end{aligned}
\end{equation}

The fifth component $\mathcal{L}_5$ discourages large discrepancies through quadratic penalties. The squared difference between modality logits creates a progressively stronger penalty as disagreement increases, effectively promoting consensus in high-stakes predictions.

The final synergy score is:
\begin{equation}
\begin{aligned}
\mathcal{S}(y_T) = \sum_{i=1}^{5} \mathcal{L}_i(y_T).
\end{aligned}
\end{equation}

The five components work together to create a robust decision mechanism that leverages each modality's strengths while mitigating weaknesses. During token generation, we apply distribution pruning based on the probability adjusted by the synergy score:

\begin{equation}
\begin{aligned}
V_{\text{head}}(y_T) = \left\{ y_t \in \mathcal{V} \mid \rho(y_t \mid v, Q, y_{<t}) \geq \beta \cdot \left( 1 + \mathcal{S}(y_T) \right) \right\},
\end{aligned}
\end{equation}

where $\mathcal{V}$ represents all possible tokens, $\beta$ is the base threshold, and $\mathcal{S}(y_T)$ adjusts the threshold based on semantic validity. This ensures only tokens with high synergy between visual and motion evidence are retained, effectively mitigating hallucinations.

The final token probability is calculated using a softmax function over the synergy scores:

\begin{equation}
\begin{aligned}
p(y_T) = \frac{\exp(\mathcal{S}(y_T))}{\sum_{y' \in \mathcal{V}} \exp(\mathcal{S}(y'))} = \text{softmax}(\mathcal{S}(y_T)).
\end{aligned}
\end{equation}

\begin{figure}[htbp] 
    \centering
    \includegraphics[width=\linewidth]{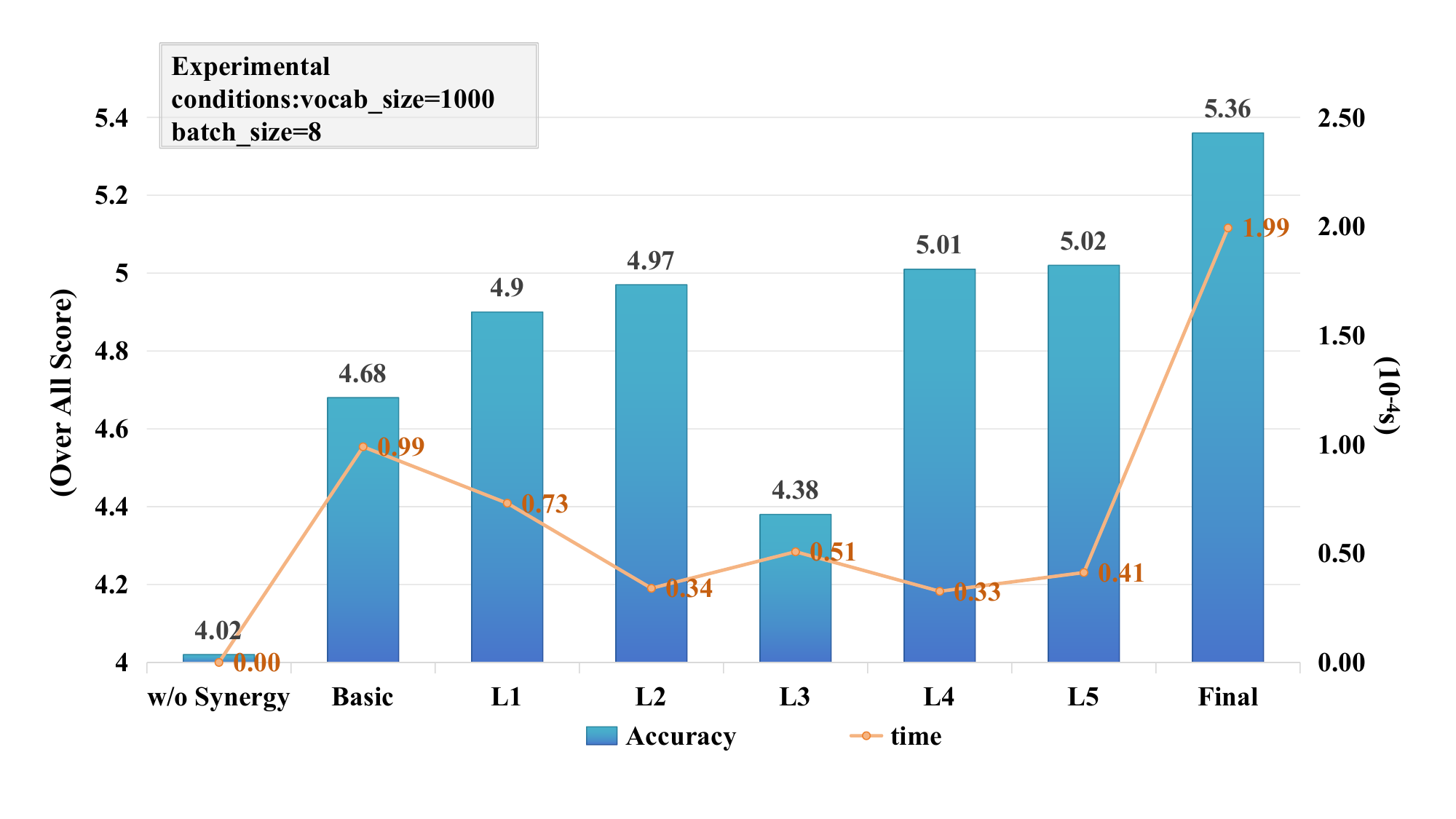} 
    \caption{Performance comparison of different synergy components in our M-ACM framework. The blue bars represent accuracy scores (left y-axis), while the orange line tracks inference time in seconds (right y-axis).}
    \label{fig:synergy_01}
    
\end{figure}

Figure~\ref{fig:synergy_01} llustrates how our different synergy components create a robust decision mechanism that strategically balances modal strengths and weaknesses. From baseline (4.02) to final combined approach (5.36), each component contributes uniquely and significantly to overall performance gains. The modest inference time increase is ultimately justified by the impressive 33.3\% accuracy improvement across diverse evaluation scenarios.

\begin{figure}[htbp] 
    \centering
    \includegraphics[width=\linewidth]{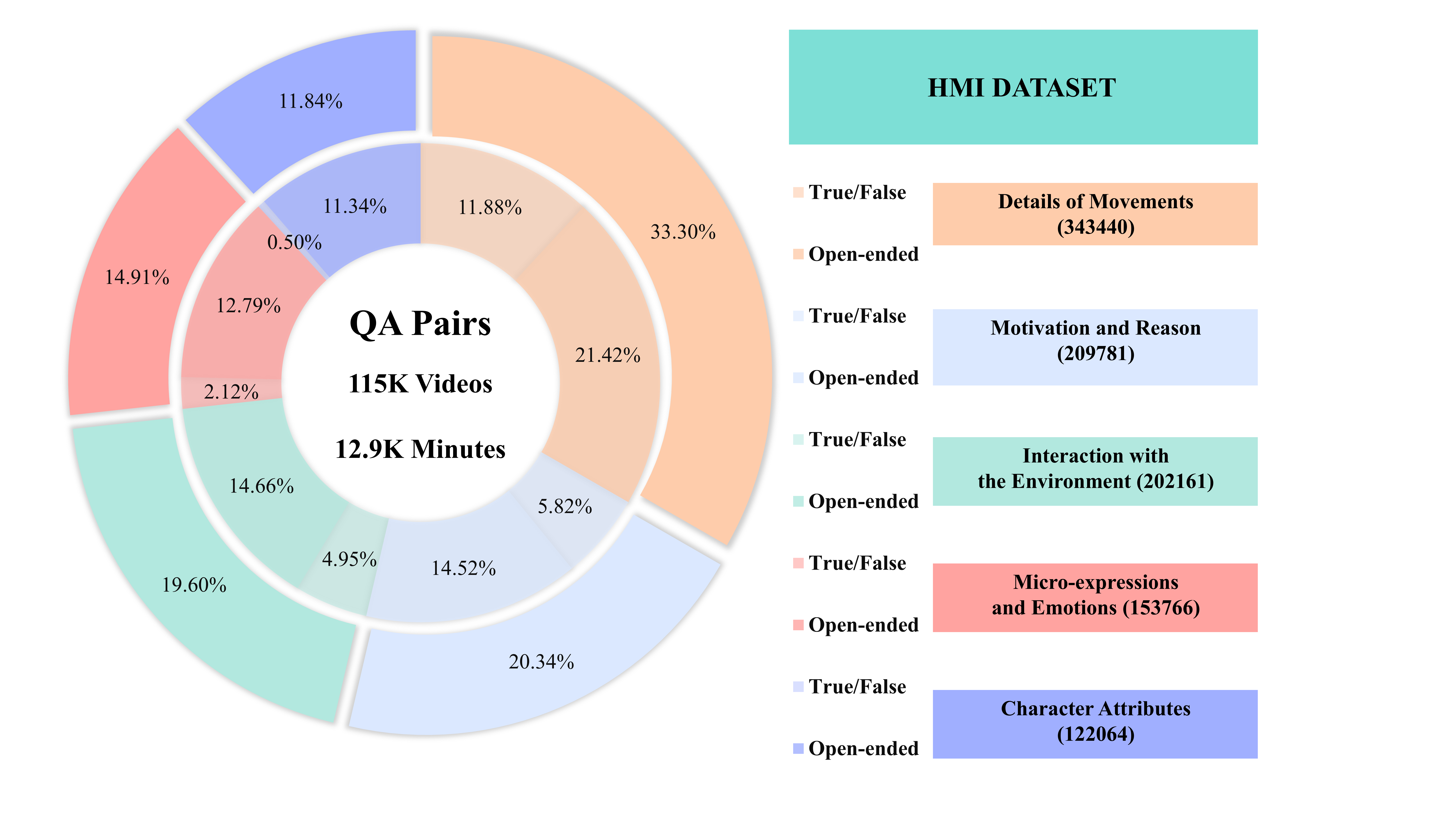} 
    \caption{The composition of QA pairs in the HMI dataset. The outer circle represents five aspects, while the inner circle shows corresponding question types for each aspect.}
    \label{fig:QA}
\end{figure}

\section{HMI Dataset}
\subsection{Overview}
\begin{figure*}[htbp] 
    \centering
    \includegraphics[width=\textwidth]{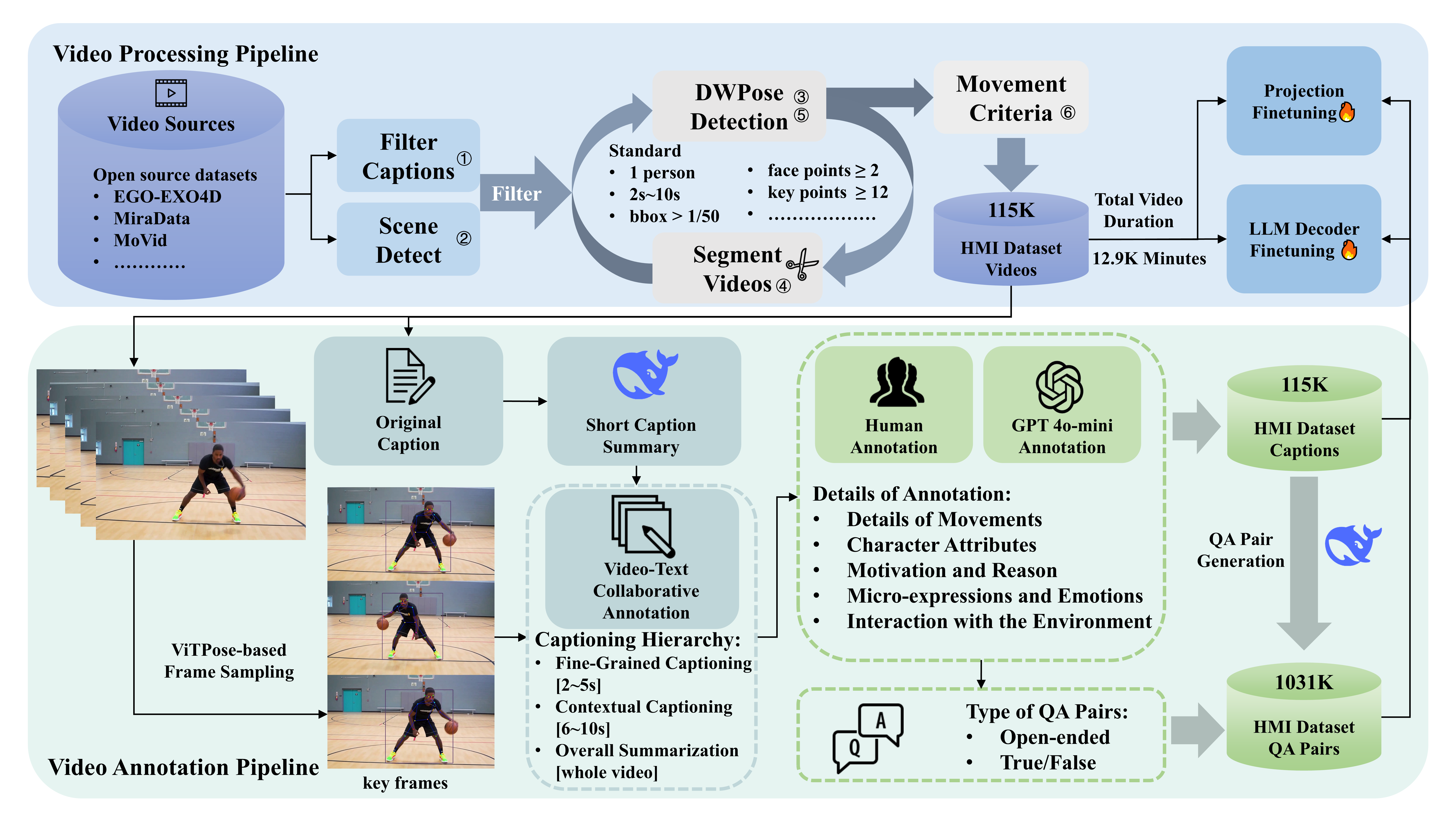} 
    \caption{The process of building the HMI dataset. Top: The video processing pipeline. Initially, the public video datasets are collected for scene segmentation and cleaning. Then, DWPose ~\cite{yang2023effective} and movement criteria are used for filtering to obtain high-quality videos. Bottom: The video annotation pipeline. Video keyframes sampled by ViTPose ~\cite{xu2022vitpose} and the original captions are utilized for video-text collaborative annotation with GPT-4o mini. Additionally, DeepSeek-R1-Distill-Qwen-7B ~\cite{guo2025deepseek} generates question-answer pairs based on the annotated video captions.}
    \label{fig:Video Processing & Annotation}
\end{figure*}

To tackle the challenge of human motion understanding, we introduce HMI, a large-scale human-centered dataset. It consists of over 115K videos, corresponding captions, and 1031K QA pairs focused on human motion, bridging the gap between video-text modalities in the domain of human motion. Figure ~\ref{fig:QA} exhibits the distribution of QA pairs for the HMI dataset, and Table ~\ref{tab:dataset} illustrates the composition of the various sub-datasets within the HMI dataset. As depicted in Figure ~\ref{fig:Video Processing & Annotation}, the construction of the HMI dataset follows a systematic process: (1) video data collection and preprocessing, (2) video filtering based on DWPose ~\cite{yang2023effective}, (3) motion quality filtering, (4) ViTPose-based ~\cite{xu2022vitpose} frame sampling, and (5) video-text collaborative annotation.

\begin{table}
  \caption{Composition of HMI Sub-Datasets. The HMI dataset is the result of processing the original five datasets.}
  \label{tab:dataset}
  \begin{tabular}{ccc}
    \toprule
    \textbf{Dataset} & \textbf{Video Count} & \textbf{Total length (minutes)} \\
    \midrule
    LLaVA-Video-178K ~\cite{zhang2024videoinstructiontuningsynthetic} & 44.2K & 4K \\
    Ego-Exo4D ~\cite{grauman2024egoexo4dunderstandingskilledhuman} & 27.6K & 4K \\
    MiraData ~\cite{ju2024miradatalargescalevideodataset} & 21.6K & 2.4K \\
    MoVid ~\cite{chen2024motionllm} & 19.6K & 2.2K \\
    ShareGPT4Video ~\cite{chen2024sharegpt4video} & 2.1K & 0.3K \\
    \midrule  
    \textbf{HMI (Ours)} & \textbf{115K} & \textbf{12.9K} \\
    \bottomrule
  \end{tabular}
\end{table}

\subsection{Video Processing Pipeline}

\subsubsection{Data Collection and Preprocessing}

The videos for the HMI dataset are collected from five datasets: MoVid \cite{chen2024motionllm} and Ego-Exo4D \cite{grauman2024egoexo4dunderstandingskilledhuman}, which focus on human motion, LLaVA-Video-178K \cite{zhang2024videoinstructiontuningsynthetic}, MiraData ~\cite{ju2024miradatalargescalevideodataset}, and ShareGPT4Video ~\cite{chen2024sharegpt4video}, which are open-source datasets used for training large language models. These sources encompass a broad range of human movements, attributes, and diverse interaction scenarios, ensuring the HMI dataset's comprehensiveness. 



Based on the original text annotations of these five datasets, we filter the videos related to human subjects using people-related keywords (e.g., "people," "man," "woman"). To maintain consistency in actions and scenes, scene transition detection is employed to segment the videos further, removing any video clips shorter than two seconds after segmentation.


\subsubsection{Video Filtering Based on DWPose}



To better capture high-quality single-person motion videos with visible human subjects, we use DWPose ~\cite{yang2023effective} for human pose estimation. Initially, over 90\% of the video frames must contain a single person, with the average bounding box area exceeding 1/50 of the image area to ensure visibility. High-quality frames must have at least 12 keypoints with confidence greater than 0.5, including 2 face keypoints, and more than 70\% of the frames must meet these criteria to ensure visibility of multiple body parts in the videos.


The videos longer than 10 seconds are split into sub-segments, each no longer than 10 seconds, with segments shorter than 2 seconds discarded. After segmentation, DWPose ~\cite{yang2023effective} is applied again using the same criteria to perform filtering.

\subsubsection{Motion Quality Filtering}

We assess motion quality by calculating the Euclidean distance of keypoints between frames. According to the OpenPose ~\cite{hidalgo2019single} format with 18 keypoints, the keypoints with confidence greater than 0.5 are normalized based on the character's bounding box and categorized into face, arms, torso, and legs. A body part is considered visible if at least two keypoints are visible. For frames where all four body parts are visible, two parts need to meet the movement criterion, while for three visible parts, all must meet the criterion. At least 50\% of frames must meet the movement condition to ensure overall motion quality.

\subsection{Video Annotation Pipeline}

\subsubsection{ViTPose-based Frame Sampling}


Text annotation of videos with multimodal LLMs is gaining popularity, but these models are limited by the number of image tokens they can process. Therefore, selecting high-quality video frames is crucial for better video understanding and text annotation. Unlike traditional uniform sampling, we propose that character motion videos can leverage the variation in human keypoints to achieve superior frame sampling results.

We extract frames containing a single person and divide them into \( N \) segments, where \( N \) is the duration of the video in seconds. In the first segment, we select the frame with the largest sum of Euclidean distances between the keypoints of all frames within this segment as the first keyframe. In the subsequent \( n \)-th segment, the frame with the greatest keypoints difference from the keyframe in the \((n-1)\)-th segment is chosen as the \( n \)-th keyframe.

\begin{equation}
\begin{aligned}
j_1 &= \arg \max_{i \in [1, M_1]} \sum_{i=1}^{M_1} \| \mathbf{k}_i^{(1)} - \mathbf{k}_j^{(1)} \|_2, \\
\end{aligned}
\end{equation}


\begin{equation}
\begin{aligned}
j_n &= \arg \max_{i \in [1, M_n]} \| \mathbf{k}_i^{(n)} - \mathbf{k}_{j_{n-1}}^{(n-1)} \|_2.
\end{aligned}
\end{equation}

Where \( j_n \) refers to the index of the \( n \)-th keyframe selected in the \( n \)-th segment. The \( \mathbf{k}_i^{(n)} \) represents the keypoints of the \( i \)-th frame in the \( n \)-th segment, and \( \mathbf{k}_{j_{n-1}}^{(n-1)} \) represents the keypoints of the keyframe in the \( (n-1) \)-th segment. The number of frames in the \( n \)-th segment is \( M_n \).



\subsubsection{Video-Text Collaborative Annotation}



We aim to extract key information from original captions to improve video annotations, summarizing them into concise descriptions of about 20 words using DeepSeek-R1-Distill-Qwen-7B ~\cite{guo2025deepseek}. This approach reduces the hallucinatory information from the original annotations while ensuring that critical and valid details are retained. Referring to the LLaVA-Video ~\cite{zhang2024videoinstructiontuningsynthetic} for annotating long videos, we adopt a hierarchical annotation strategy for the videos.



As depicted in Figure ~\ref{fig:Video Processing & Annotation}, we utilize DeepSeek ~\cite{guo2025deepseek} to design QA pairs from different aspects based on the annotations of the videos, enabling the model to better understand the character's actions from multiple perspectives. The QA pairs constructed for the HMI dataset encompass five distinct perspectives, each featuring two types of questions: open-ended and True/False. Figure ~\ref{fig:QA} illustrates the distribution of the different aspects and types of QA pairs.

\subsection{HMI-Bench}

Current benchmarks for human video understanding primarily focus on the overall motion, without delving into the intricate details of human motion. To address this gap, we introduce HMI-Bench, designed to better assess the ability of multimodal LLMs to understand character motion. HMI-Bench consists of videos from MotionBench ~\cite{hong2024motionbench} and Ego-Exo4D ~\cite{grauman2024egoexo4dunderstandingskilledhuman}, totaling approximately 1.1K videos, along with their corresponding video captions and QA pairs. Of these, The 3.3K QA pairs includes about 2.1K open-ended questions related to the details of movements and 1.2K True/False questions. It is important to note that the videos from Ego-Exo4D ~\cite{grauman2024egoexo4dunderstandingskilledhuman} in HMI-Bench do not overlap with those in the HMI dataset. These videos have been processed through the video processing and annotation pipelines, and manually reviewed by humans.

\begin{table*}
  \caption{Performance comparison of different models on standard caption metrics in HMI-Bench.}
  \label{tab:caption_metrics}
  \begin{tabular}{lccccccccc}
    \toprule
    \textbf{Model} & \textbf{BLEU-1} & \textbf{BLEU-2} & \textbf{BLEU-3} & \textbf{BLEU-4} & \textbf{METEOR} & \textbf{ROUGE-1} & \textbf{ROUGE-2} & \textbf{ROUGE-L} & \textbf{CIDEr} \\
    \midrule
    MotionLLM ~\cite{chen2024motionllm}& 0.01443 & 0.00647 & 0.00242 & 0.00114 & 0.05923 & 0.14352 & 0.02663 & 0.13150 & 0.19101 \\
    VideoLLaVA ~\cite{videollava}& 0.01539 & 0.00827 & 0.00353 & 0.00175 & 0.07488 & 0.17619 & 0.05160 & 0.16774 & 0.30844 \\
    LLaVAOV-Qwen2 0.5B ~\cite{LLaVAOneVision2024}& 0.00317 & 0.00152 & 0.00066 & 0.00034 & 0.05186 & 0.15367 & 0.02469 & 0.13830 & 0.28870 \\
    LLaVAOV-Llama3 8B  ~\cite{LLaVAOneVision2024}& 0.20638 & 0.09195 & 0.03969 & 0.01883 & 0.16258 & 0.21745 & 0.04641 & 0.19562 & 0.33072 \\
    LLaVAOV-Qwen2 7B  ~\cite{LLaVAOneVision2024}& 0.32558 & 0.14663 & 0.06809 & 0.03311 & 0.22299 & 0.23203 & 0.05359 & 0.20679 & 0.33410 \\
    QwenVL2-7B  ~\cite{Qwen2VL2024}& 0.00415 & 0.00236 & 0.00118 & 0.00056 & 0.04983 & 0.14249 & 0.04469 & 0.13665 & 0.38995 \\
    QwenVL2.5-7B ~\cite{bai2025qwen2.5vl} & 0.15261 & 0.07716 & 0.03853 & 0.02026 & 0.16017 & 0.24990 & 0.06449 & 0.22859 & 0.55343 \\
    \midrule
    M-ACM Llama3 8B (Ours) & 0.49301 & 0.29403 & 0.18370 & 0.11740 & 0.33590 & 0.34757 & 0.13670 & 0.31458 & 0.72163 \\
    M-ACM Qwen2 7B (Ours)& \textbf{0.49520} & \textbf{0.29957} & \textbf{0.19070} & \textbf{0.12421} & \textbf{0.33775} & \textbf{0.36375} & \textbf{0.14496} & \textbf{0.32737} & \textbf{0.80662} \\
    \bottomrule
  \end{tabular}
\end{table*}

\section{Experiments}

\subsection{Experiment Setup}

\subsubsection{Models and Datasets}

We implemented our approach on two foundation models: Qwen2 7B ~\cite{yang2024qwen2technicalreport} (with SigLip ~\cite{zhai2023sigmoid} vision encoder) and Llama3 8B ~\cite{grattafiori2024llama} (with CLIP ~\cite{Learning2021} vision encoder). Both models were fine-tuned on the Human Motion Insight (HMI) dataset, which contains over 115K video clips with fine-grained human motion descriptions. For training, we used the complete caption dataset (115K samples) and 20\% of the QA pairs (approximately 206K samples) to ensure comprehensive coverage of both descriptive and interactive aspects of human motion understanding.

\subsubsection{Evaluation Metrics}

For a comprehensive assessment, we employed standard captioning metrics including BLEU-1/2/3/4 ~\cite{papineni2002bleu}, METEOR ~\cite{banerjee2005meteor}, ROUGE-1/2/L ~\cite{lin2004rouge}, and CIDEr ~\cite{vedantam2015cider} to assess general caption quality. Additionally, we developed a GPT-4-based evaluation framework that scores outputs across five dimensions: Details of Movements, Interaction with Environment, Motivation and Reason, Micro-expressions and Emotions, Character Attributes, and Overall Score  
 calculated as the average scores across all dimensions, which provides a holistic evaluation of the output's quality.



\subsubsection{Baseline Algorithms and Implementation Details}

We compared our approach against several state-of-the-art vision-language models including: VideoLLaVA ~\cite{videollava}, MotionLLM ~\cite{chen2024motionllm}, Qwen2-VL-7B \cite{Qwen2VL2024}, Qwen2.5-VL-7B \cite{bai2025qwen2.5vl}, and multiple versions of LLaVA-OV \cite{LLaVAOneVision2024} spanning different parameter scales (Qwen2 0.5B, Qwen2 7B ~\cite{yang2024qwen2technicalreport}, Llama3 8B ~\cite{grattafiori2024llama}). These baselines represent the current state-of-the-art in video understanding and motion description tasks.

The fine-tuning of M-ACM was conducted on 8 A100 GPUs, requiring approximately 35 hours for the Qwen2 7B ~\cite{yang2024qwen2technicalreport} model and comparable time for the Llama3 8B ~\cite{grattafiori2024llama} model. Empirically, we set the hyperparameters \( \alpha_1 \) through \( \alpha_6 \) to 0.5, 0.2, 0.4, 0.8, 0.3, and 2.0 respectively in our synergy calculation implementation. And the distribution pruning threshold \( \beta \) was set to 0.2, providing optimal balance between preserving valid tokens and removing potentially hallucinated content.

\begin{table*}
  \caption{Performance comparison of different models on video captioning in HMI-Bench across five dimensions.}
  \label{tab:motion_understanding}
  \begin{tabular}{lcccccc}
    \toprule
    \textbf{Model} & \textbf{Details of} & \textbf{Interaction with} & \textbf{Motivation} & \textbf{Micro-expressions} & \textbf{Character} & \textbf{Overall} \\
    & \textbf{Movements} & \textbf{Environment} & \textbf{and Reason} & \textbf{and Emotions} & \textbf{Attributes} & \textbf{Score} \\
    \midrule
    MotionLLM ~\cite{chen2024motionllm}& 1.40 & 1.67 & 0.98 & 0.30 & 1.77 & 1.40 \\
    VideoLLaVA ~\cite{videollava}& 1.96 & 2.58 & 0.87 & 0.23 & 2.70 & 1.67 \\
    LLAVAOV-Qwen2 0.5B ~\cite{LLaVAOneVision2024} & 2.48 & 2.87 & 1.39 & 0.53 & 2.54 & 1.96 \\
    LLAVAOV-Llama3 8B ~\cite{LLaVAOneVision2024}& 3.62 & 5.28 & 2.97 & 1.28 & 4.82 & 3.62 \\
    LLAVAOV-Qwen2 7B ~\cite{LLaVAOneVision2024}& 3.83 & 6.22 & 2.03 & 1.28 & 6.00 & 3.83 \\
    QwenVL2 7B  ~\cite{Qwen2VL2024}& 1.82 & 2.72 & 1.05 & 0.34 & 2.74 & 1.82 \\
    QwenVL2.5 7B  ~\cite{bai2025qwen2.5vl}& 4.42 & \textbf{6.44} & 3.60 & 1.54 & 5.40 & 4.42 \\
    \midrule
    M-ACM Llama3 8B (Ours) & 4.36 & 4.34 & 4.01 & 4.81 & 4.56 & 4.42 \\
    M-ACM Qwen2 7B (Ours) & \textbf{5.61} & 5.91 & \textbf{5.05} & \textbf{5.54} & \textbf{6.32} & \textbf{5.69} \\
    \bottomrule
  \end{tabular}
\end{table*}

\begin{table*}[htbp]
\centering
\caption{Ablation studies on video captioning in HMI-Bench from five evaluation perspectives.}
\label{tab:HMI video caption}
\begin{tabular}{lcccccccccc}
\toprule
\textbf{Model} & \textbf{MSD} & \textbf{HMI Dataset} & \textbf{ViTPose ~\cite{xu2022vitpose}} & \textbf{DM} & \textbf{IE} & \textbf{MR} & \textbf{ME} & \textbf{CA} & \textbf{Overall Score} \\
\midrule
LLAVAOV-Qwen2 7B ~\cite{LLaVAOneVision2024} & $ \times $ & $ \times $ & $ \times $ & 3.62 & 6.22 & 2.03 & 1.28 & 6.00 &  3.83 \\
LLAVAOV-Qwen2 7B ~\cite{LLaVAOneVision2024} & $ \times $ & $ \checkmark $ & $ \times $ & 5.20 & 5.81 & 4.73 & 5.22 & 5.61 & 5.31 \\
M-ACM Qwen2 7B & $ \checkmark $ & $ \times $ & $ \times $ & 3.59 & 6.15 & 2.12 & 1.47 & 5.94 & 3.85 \\
M-ACM Qwen2 7B & $ \checkmark $ & $ \checkmark $ & $ \times $ & 5.51 & 5.81 & 4.92 & 5.36 & 6.03 &  5.53 \\
M-ACM Qwen2 7B & $ \checkmark $ & $ \checkmark $ & $ \checkmark $ & 5.61 & 5.91 & 5.05 & 5.54 & 6.32 & 5.69 \\
\bottomrule
\end{tabular}
\end{table*}

\begin{table*}[htbp]
\centering
\caption{Ablation studies on QA tasks in HMI-Bench with open-ended and True/False types.}
\label{tab:HMI-QA}
\begin{tabular}{lccc ccccc}
\toprule
\textbf{Model} & \textbf{MSD} & \textbf{HMI Dataset} & \textbf{ViTPose ~\cite{xu2022vitpose}} & \textbf{DM} & \multicolumn{4}{c}{\textbf{True/False}} \\
\cmidrule(lr){6-9}
& & & & & \textbf{Accuracy} & \textbf{Precision} & \textbf{Recall} & \textbf{F1 Score} \\
\midrule
LLAVAOV-Qwen2 7B ~\cite{LLaVAOneVision2024} & $ \times $ & $ \times $ & $ \times $ & 2.7972 & 0.5461 & 0.6782 & 0.3496 & 0.4614 \\
LLAVAOV-Qwen2 7B ~\cite{LLaVAOneVision2024} & $ \times $ & $ \checkmark $ & $ \times $ & 5.7099 & 0.7377 & 0.7139 & 0.8822 & 0.7892 \\
M-ACM Qwen2 7B & $ \checkmark $ & $ \times $ & $ \times $ & 2.8382 & 0.5539 & 0.7209 & 0.3215 & 0.4447 \\
M-ACM Qwen2 7B & $ \checkmark $ & $ \checkmark $ & $ \times $ & 5.8483 & 0.7795 & 0.7458 & 0.9161 & 0.8222 \\
M-ACM Qwen2 7B & $ \checkmark $ & $ \checkmark $ & $ \checkmark $ & 5.8533 & 0.7787 & 0.7455 & 0.9146 & 0.8214 \\
\bottomrule
\end{tabular}
\end{table*}

\subsection{Main Results}

Table \ref{tab:caption_metrics} presents performance comparisons across standard caption metrics. The M-ACM Qwen2 7B model achieves the best performance across all metrics, with particularly notable gains in BLEU-4 ~\cite{papineni2002bleu} (0.12421 vs. 0.03311 for LLAVAOV-Qwen2 7B ~\cite{LLaVAOneVision2024}) and CIDEr ~\cite{vedantam2015cider} (0.80662 vs. 0.55343 for QwenVL2.5-7B ~\cite{bai2025qwen2.5vl}). This represents approximately a 3.7× improvement in BLEU-4 ~\cite{papineni2002bleu} and a 1.5× improvement in CIDEr ~\cite{vedantam2015cider} compared to the best baseline. These significant gains can be attributed to our MSD approach, and the consistent improvements across all metrics highlight the robustness of our approach in capturing fine-grained motion details.

\begin{figure}[htbp] 
    \centering
    \includegraphics[width=\linewidth]{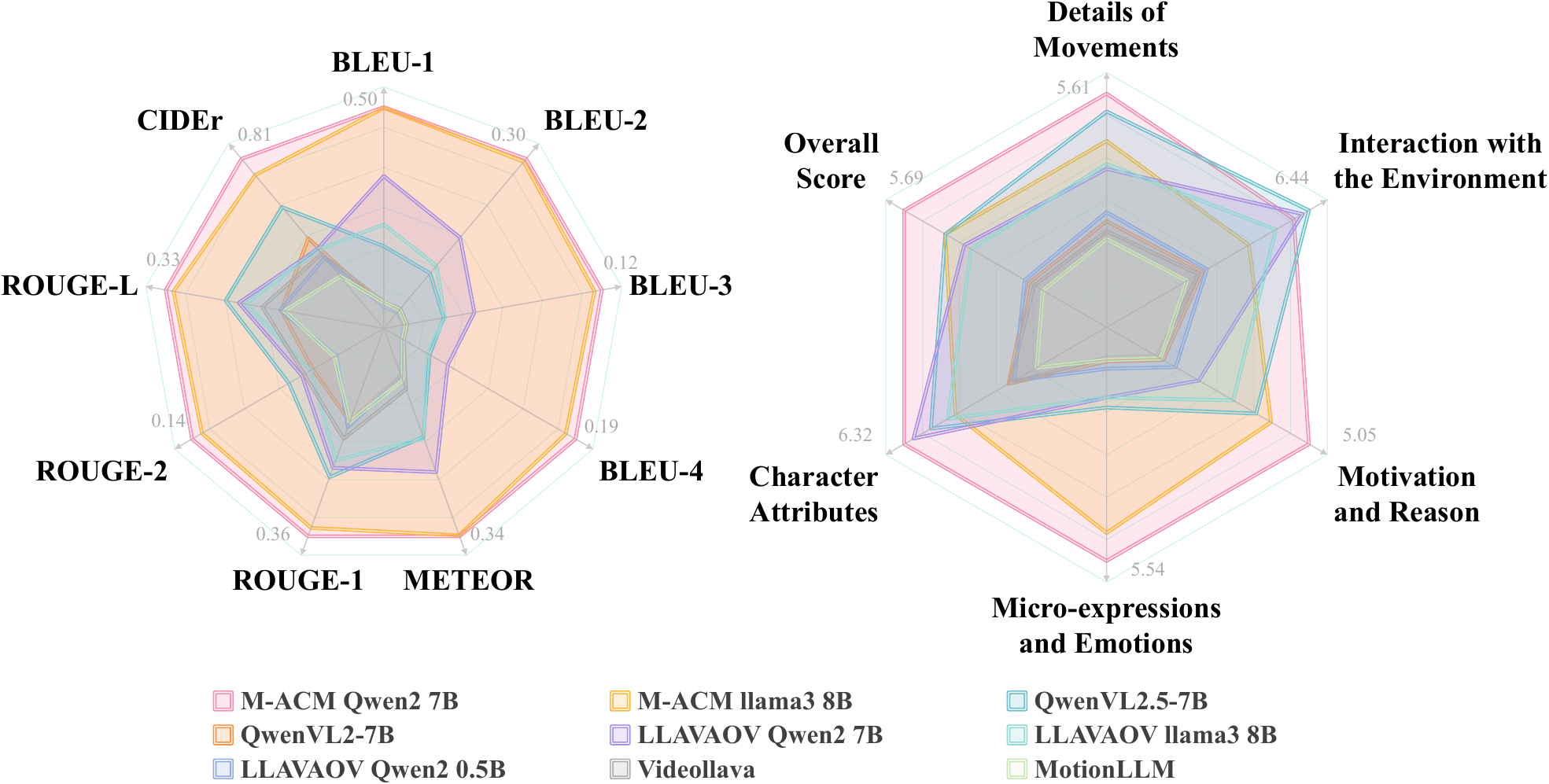} 
    \caption{Performance comparison of M-ACM against other models.The radar plots illustrate our model's superior performance across standard caption metrics (left) and human motion understanding dimensions (right).}
    \label{fig:radar}
\end{figure}

Table \ref{tab:motion_understanding} shows performance comparisons on human motion understanding across five key dimensions. The results clearly demonstrate the superiority of our M-ACM , particularly in aspects requiring fine-grained motion understanding. The M-ACM Qwen2 7B model outperforms all baselines in four of the five dimensions, with particularly substantial improvements in "Motivation and Reason" (5.05 vs. 3.60 for QwenVL2.5-7B ~\cite{bai2025qwen2.5vl}) and "Micro-expressions and Emotions" (5.54 vs. 1.54 for QwenVL2.5-7B ~\cite{bai2025qwen2.5vl}). This represents a 40\% improvement in understanding motivations and a 260\% improvement in detecting subtle emotions. 

Figure ~\ref{fig:radar} presents a comprehensive visualization of our evaluation results. The substantial area coverage of our M-ACM in both radar charts confirms its advantages over existing approaches. While traditional metrics show our numerical superiority, the human motion understanding dimensions reveal our model's true strength - capturing nuanced movements, emotional states, and contextual reasoning that eludes conventional vision-language systems. This performance gap is particularly pronounced in challenging aspects like micro-expressions and motivation understanding, where our synergetic decoding mechanism effectively leverages complementary information to overcome the hallucination problems that plague previous methods.

\subsection{Ablation Studies}

In order to validate the effectiveness of ViTPose-based keyframe sampling in the M-ACM, as well as the improvement brought by the HMI dataset and the MSD method in understanding human motion, we choose Qwen2 as the base model and conduct several ablation experiments on both video captions and QA tasks within the HMI-Bench.

As shown in the Table ~\ref{tab:HMI video caption} and ~\ref{tab:HMI-QA}, the introduction of the HMI dataset improves the overall video caption score by 38.6\%, the motion detail score in the QA by 104\%, and the accuracy of the judgment questions by 35\%.  The MSD method consistently enhances the M-ACM's ability to understand human motion in both video captioning and QA. Moreover, the incorporation of the keyframe sampling, built upon the first two improvements, results in an additional 0.16 increase in the video captioning score in HMI-Bench. Ultimately, our M-ACM Qwen2 7B outperforms LLAVAOV-Qwen2 7B by 48\% in video captioning, and achieves a 109\% improvement in the motion detail score and a 42.6\% increase in accuracy for both open-ended and True/False types in the QA tasks.

\subsection{Further Discussions} 
Our M-ACM framework significantly reduces hallucinations and enhances motion caption accuracy via synergistic dual-path integration, enabling precise interpretation of subtle movements missed by conventional models. Future work will focus on enhancing motion representation by annotating SMPL format data using our HMI dataset to train a dedicated motion encoder.

\section{Conclusion}

We present the  Motion-Augmented Caption Model (M-ACM) with Motion Synergetic Decoding (MSD), a novel dual-pathway architecture that bridges visual understanding and motion-specific representation for accurate human motion captioning. To bridge the gap between the video-text modalities in the human motion domain, we construct the Human Motion Interpretation (HMI) dataset, containing 115K videos, 115K captions, and 1031K QA pairs, which enhances M-ACM’s comprehensive understanding into human motion. By calculating synergy between modalities and integrating SMPL-based motion representations, our M-ACM produces precise movement descriptions with reduced hallucination issues. Experiments on HMI-Bench demonstrate that M-ACM substantially outperforms existing methods across both standard captioning metrics and specialized motion understanding dimensions.

\begin{acks}
This research was funded through National Key Research and Development Program of China (Project No. 2022YFB36066),in part by the Shenzhen Science and Technology Project under Grant (KJZD20240903103210014,JCYJ20220818101001004).
\end{acks}

\bibliographystyle{ACM-Reference-Format}
\balance
\bibliography{sigconf_msd}










\end{document}